\documentclass[runningheads]{llncs}
\usepackage[T1]{fontenc}
\usepackage{graphicx}
\usepackage{microtype}
\usepackage{booktabs}
\usepackage{multirow}
\usepackage{svg}
\usepackage{amsmath,amssymb}
\usepackage{here}
\usepackage{url}
\usepackage{comment}
\usepackage{multirow}
\usepackage{tabularx}
\usepackage{color}
\usepackage{soul}
\usepackage{tcolorbox}
\usepackage{colortbl}

\newcounter{num}
\newcommand{\Rnum}[1]{\setcounter{num}{#1}\Roman{num}}

\begin{document}
\title{Improving the Robustness to Variations of Objects and Instructions with a Neuro-Symbolic Approach for Interactive Instruction Following}
%
%

\author{Kazutoshi Shinoda\inst{1,2} \and Yuki Takezawa\inst{3} \and Masahiro Suzuki\inst{1} \and Yusuke Iwasawa\inst{1} \and Yutaka Matsuo\inst{1}}

\authorrunning{K. Shinoda et al.}
\titlerunning{Improving the Robustness to Variations of Objects and Instructions}

%

\institute{The University of Tokyo, Tokyo, Japan\\
\and
National Institute of Informatics, Tokyo, Japan\\
\and
Kyoto University, Kyoto, Japan\\
\email{shinoda@is.s.u-tokyo.ac.jp}
}

\maketitle              
\begin{abstract}
An interactive instruction following task has been proposed as a benchmark for learning to map natural language instructions and first-person vision into sequences of actions to interact with objects in 3D environments.
We found that an existing end-to-end neural model for this task tends to fail to interact with objects of unseen attributes and follow various instructions.
We assume that this problem is caused by the high sensitivity of neural feature extraction to small changes in vision and language inputs.
To mitigate this problem, we propose a neuro-symbolic approach that utilizes high-level symbolic features, which are robust to small changes in raw inputs, as intermediate representations.
We verify the effectiveness of our model with the subtask evaluation on the ALFRED benchmark.
Our experiments show that our approach significantly outperforms the end-to-end neural model by 9, 46, and 74 points in the success rate on the ToggleObject, PickupObject, and SliceObject subtasks in unseen environments respectively.

\end{abstract}

\section{Introduction}
To operate robots in human spaces, instruction following tasks in 3D environments have attracted substantial attention \cite{Anderson_2018_CVPR,Chen_2019_CVPR,puig-etal-2018-virtualhome}.
In these tasks, robots are required to translate natural language instructions and egocentric vision into sequences of actions.
To enable robots to perform further complex tasks that require interaction with objects in 3D environments, the ``interactive instruction following'' task has been proposed \cite{ALFRED20}.
Here, interaction with objects refers to the movement or the state change of objects caused by actions such as picking up or cutting.

\begin{figure}[tbp]
    \centering
    \begin{tabular}{cccc}
        \begin{minipage}{25mm}
            \centering
          \scalebox{0.25}{\includegraphics{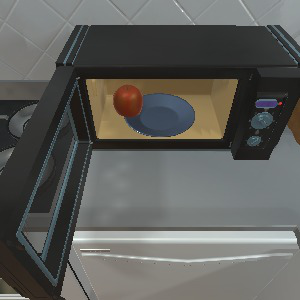}}
        \end{minipage} & \hspace{2mm}
        \begin{minipage}{25mm}
            \centering
        \scalebox{0.25}{\includegraphics{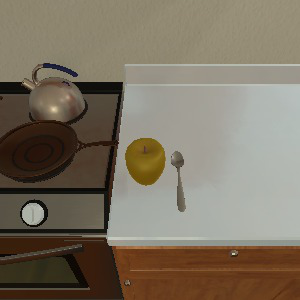}}
        \end{minipage} & \hspace{2mm}
        \begin{minipage}{25mm}
            \centering
          \scalebox{0.25}{\includegraphics{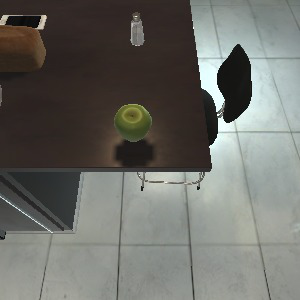}}
        \end{minipage} & \hspace{2mm}
        \begin{minipage}{25mm}
            \centering
        \scalebox{0.25}{\includegraphics{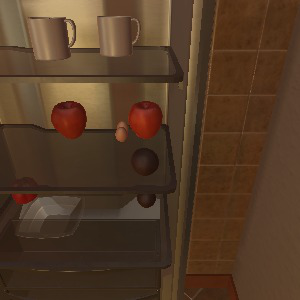}}
        \end{minipage}\\
    \end{tabular}
    \caption{An example of four different apples that an agent is required to pick up, taken from the ALFRED benchmark \cite{ALFRED20}. An agent is required to interact with objects of various shapes, colors, and textures.}
    \label{fig:apple}
\end{figure}

\begin{figure}[htbp]
    \centering
    \scalebox{0.3}{\includegraphics{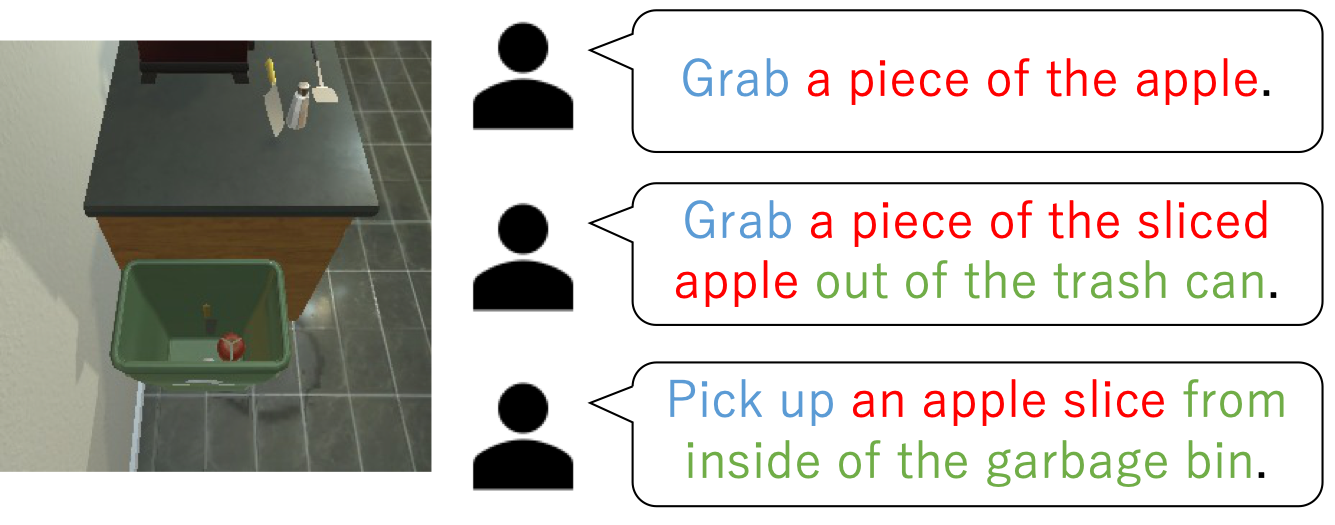}}
    \caption{An example where different language instructions are given by different annotators to the same action, taken from the ALFRED benchmark \cite{ALFRED20}.
    Predicates (blue), referring expressions (red), and modifiers (green) have the same meaning but can be expressed in various ways. Modifiers can be omitted.
    Agents should take the correct action consistently no matter how the given instruction is expressed.
    }
    \label{fig:pickup}
\end{figure}

In interactive instruction following, agents need to be robust to variations of objects and language instructions that are not seen during training.
For example, as shown in Figure \ref{fig:apple}, objects are of the same class but vary in attributes such as color, shape, and texture.
Also, as shown in Figure \ref{fig:pickup}, language instructions vary in predicates, referring expressions pointing to objects, and the presence or absence of modifiers, even though their intents are the same.

However, our analysis revealed that the end-to-end neural baseline proposed by Shridhar et al. \cite{ALFRED20} for the task is not robust to variations of objects and language instructions, i.e., it often fails to interact with objects of unseen attributes or to take the correct actions consistently when language instructions are replaced by their paraphrases.
Similar phenomena have been observed in the existing studies.
For example, end-to-end neural models that compute outputs from vision or language inputs with only continuous representations in the process are shown to be sensitive to small perturbations in inputs in image classification \cite{2013arXiv1312.6199S} and natural language understanding \cite{jia-liang-2017-adversarial}.

Given these observations, we hypothesize that reasoning over the high-level symbolic representations of objects and language instructions are robust to small changes in inputs.
In this study, we aim to mitigate this problem by utilizing high-level symbolic representations that can be extracted from raw inputs and reasoning over them.
Specifically, high-level symbolic representations in this study refer to classes of objects, high-level actions, and their arguments of language instructions.
These symbolic representations are expected to be robust to small changes in the input because of their discrete nature.

Our contributions are as follows.
\begin{itemize}
    \vspace{-8pt}
    \item We propose Neuro-Symbolic Instruction Follower (NS-IF), which introduces high-level symbolic feature extraction and reasoning modules to improve the robustness to variations of objects and language instructions for the interactive instruction following task.
    \item In subtasks requiring interaction with objects, our NS-IF significantly outperforms an existing end-to-end neural model, S2S+PM, in the success rate while improving the robustness to the variations of vision and language inputs.
\end{itemize}

\section{Neuro-Symbolic Instruction Follower}
\begin{figure}[t]
\centering
\includegraphics[width=9cm]{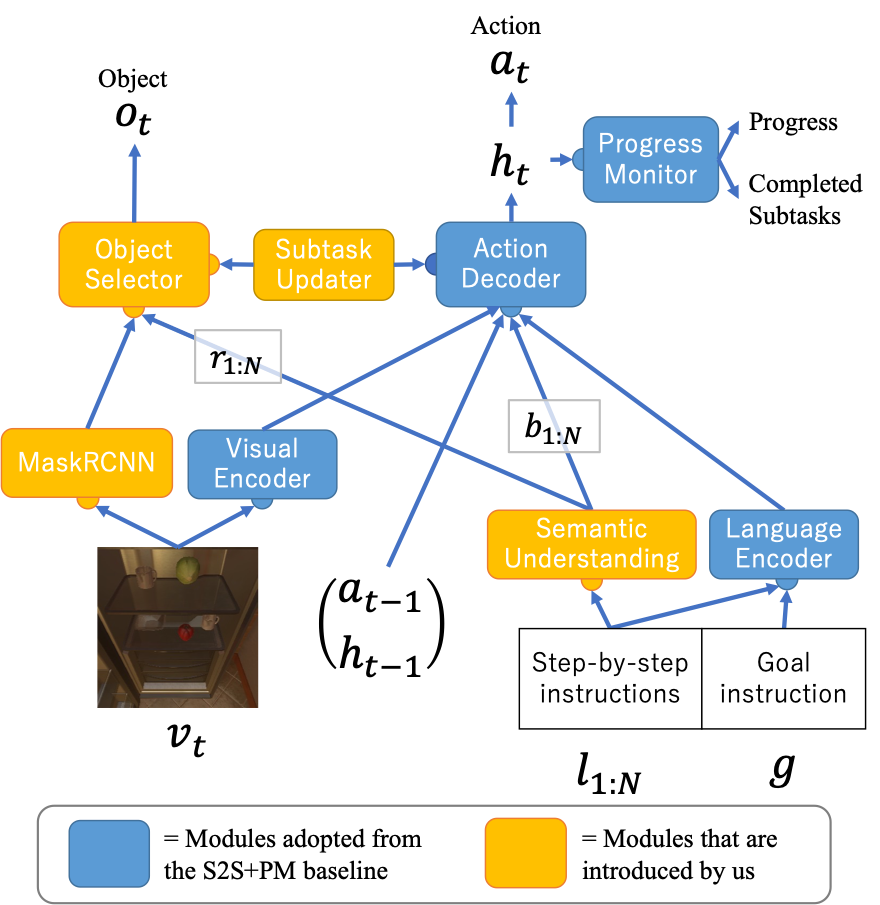}
\caption{Overview of the proposed NS-IF. The modules are colored to clarify the difference between the S2S+PM baseline \cite{ALFRED20} and our NS-IF.}
\label{fig:overview}
\end{figure}

We propose Neuro-Symbolic Instruction Follower (NS-IF) to improve the robustness to variations of objects and language instructions as illustrated in Figures \ref{fig:apple} and \ref{fig:pickup}.
The whole picture of the proposed method is shown in Figure \ref{fig:overview}.
Specifically, different from the S2S+PM baseline \cite{ALFRED20}, we introduce semantic understanding module (\S\ref{sec:semantic-understanding}) and MaskRCNN (\S\ref{sec:maskrcnn}) to extract high-level symbolic features from raw inputs, subtask updater (\S\ref{sec:subtask-updater}) to make the model recognize which subtask is being solved, and object selector (\S\ref{sec:object-selector}) to make robust reasoning over the extracted symbolic features.
Other components are adopted following S2S+PM.
Each component of NS-IF is explained below in detail.

\begin{figure*}[tbp]
    \centering
    \begin{tabular}{c}
            \scalebox{0.4}{\includegraphics{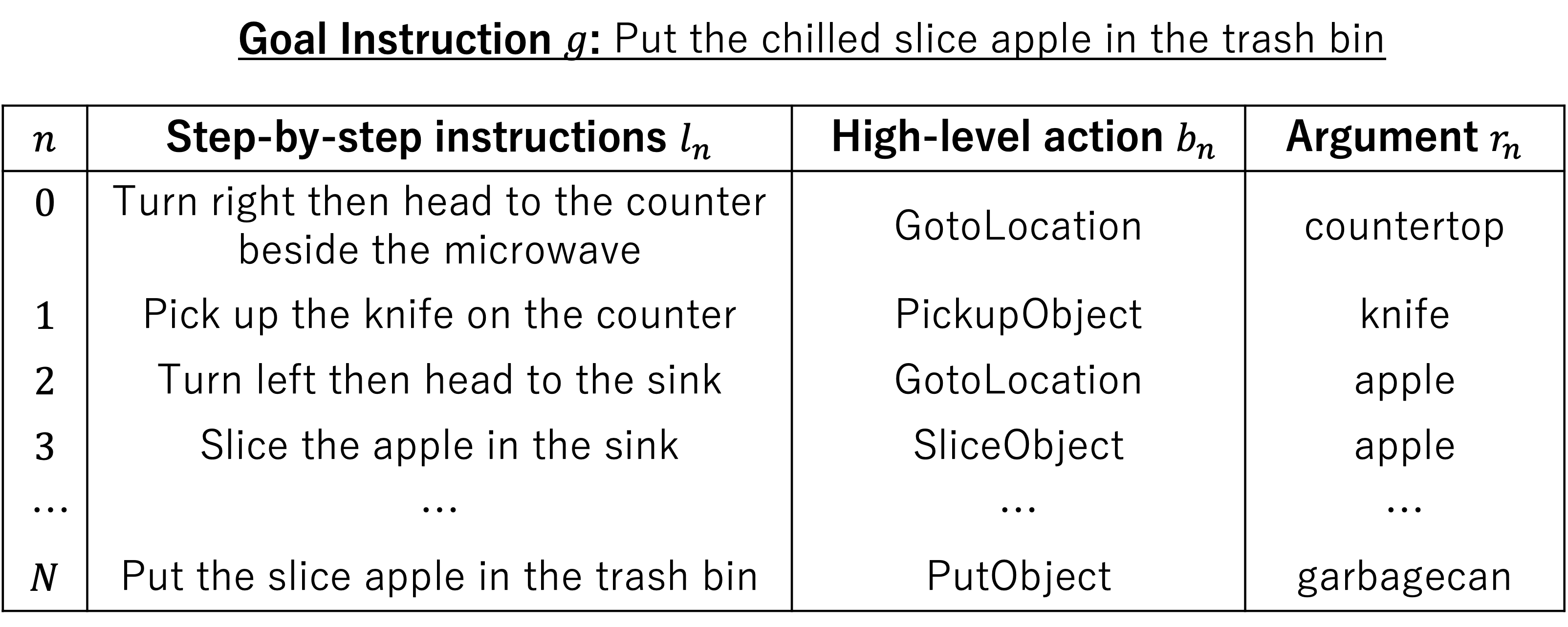}}
        \\
        (a) Instructions and their high-level actions and arguments\vspace{5pt}\\
            \scalebox{0.4}{\includegraphics{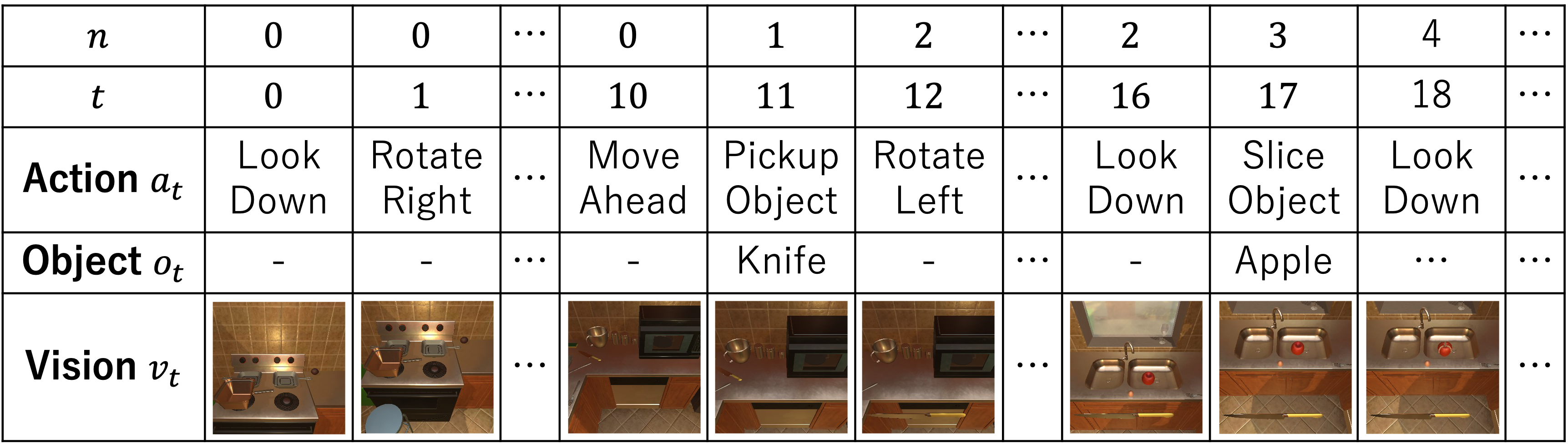}}
        \\
        (b) Visual inputs and ground-truth actions and objects for each time step\\
    \end{tabular}
    \caption{An example of the interactive instruction following task taken from ALFRED.}
    \label{fig:example}
\end{figure*}

\subsection{Notation}
The length of the sequence of actions required to accomplish a task is $T$.
The action at time $t$ is $a_t$.
The observed image at time $t$ is $v_t$.
The total number of subtasks is $N$.
The step-by-step language instruction for the $n$-th subtask is $l_n$, and the language instruction indicating the goal of the overall task is $g$.
Let $b_n$ be the high-level action for the language instruction $l_n$ for each subtask, and $r_n$ be its argument.
The total number of observable objects in $v_t$ is $M$.
The mask of the $m$-th object is $u_m$, and the class of the $m$-th object is $c_m$.
An example is displayed in Figure \ref{fig:example}.

\subsection{Language Encoder}
The high-level symbolic representations of step-by-step language instructions consist of only the high-level actions $b_{1:N}$ and the arguments $r_{1:N}$, and information about modifiers is lost.
To avoid the failure caused by the lack of information, we input all the words in the language instructions to the language encoder to obtain continuous representations.
The word embeddings of the language instruction $g$ representing the goal and the step-by-step language instruction $l_{1:N}$ for all subtasks are concatenated and inputted into bidirectional LSTM \cite{lstm} (BiLSTM) to obtain a continuous representation $H$ of the language instruction.\footnote{When using only high-level symbolic expressions as input to the BiLSTM, the accuracy decreased. Therefore, we use continuous representation as input here.}

\subsection{Visual Encoder}
Similarly, for the image $v_t$, a continuous representation $V_t$ is obtained with ResNet-18 \cite{resnet}, whose parameters are fixed during training.

\subsection{Semantic Understanding}
\label{sec:semantic-understanding}
Here, we convert the language instructions $l_n$ for each subtask into high-level actions $b_n$ and their arguments $r_n$.
To this end, we trained RoBERTa-base \cite{liu2019roberta} on the ALFRED training set.
We adopted RoBERTa-base here because it excels BERT-base \cite{bert} in natural language understanding tasks \cite{liu2019roberta}.
For predicting $b_n$ and $r_n$ from $l_n$, two classification heads are added in parallel on top of the last layer of RoBERTa-base.

We used the ground truth $b_n$ and $r_n$ provided by ALFRED during training.
At test time, we used $b_n$ and $r_n$ predicted by the RoBERTa-base.
To see the impact of the prediction error of semantic understanding, we also report the results when using the ground truth $b_n$ and $r_n$ at test time.

\subsection{MaskRCNN}
\label{sec:maskrcnn}
MaskRCNN \cite{maskrcnn} is used to obtain the masks $u_{1:M}$ and classes $c_{1:M}$ of each object from the image $v_t$.
Here, we use a MaskRCNN pre-trained on ALFRED.\footnote{https://github.com/alfworld/alfworld}

\subsection{Subtask Updater}
\label{sec:subtask-updater}
We find that the distribution of the output action sequences varies greatly depending on which subtask is being performed.
In this section, to make it easier to learn the distribution of the action sequences, the subtask $s_t$ being performed is predicted at each time.
Since our aim is to evaluate the approach on each subtask, we conducted experiments under the condition that the ground truth $s_t$ is given during both training and testing.

\subsection{Action Decoder}
The action decoder predicts the action $a_t$ at each time using LSTM.
Different from S2S+PM, the action decoder takes high-level actions $b_{1:N}$ as inputs.
Namely, the inputs are the hidden state vector $h_{t-1}$ at time $t-1$, the embedding vector of the previous action $a_{t-1}$, the embedding representation of the high-level action $E(b_{1:N})^T p(s_t)$ and $V_t$ at time $t$ obtained using the embedding layer $E$ and $s_t$, and the output $x_{t-1}$ from $h_{t-1}$ to $H$. $V_t$, and $w_t$, which is the concatenation of the output $x_t$ of attention from $h_{t-1}$ to $H$.
Then, after concatenating $w_t$ to the output $h_t$ of LSTM, we obtain the distribution of behavior $a_t$ via linear layer and Softmax function.

\begin{figure*}[tbp]
\centering
\scalebox{0.4}{\includegraphics{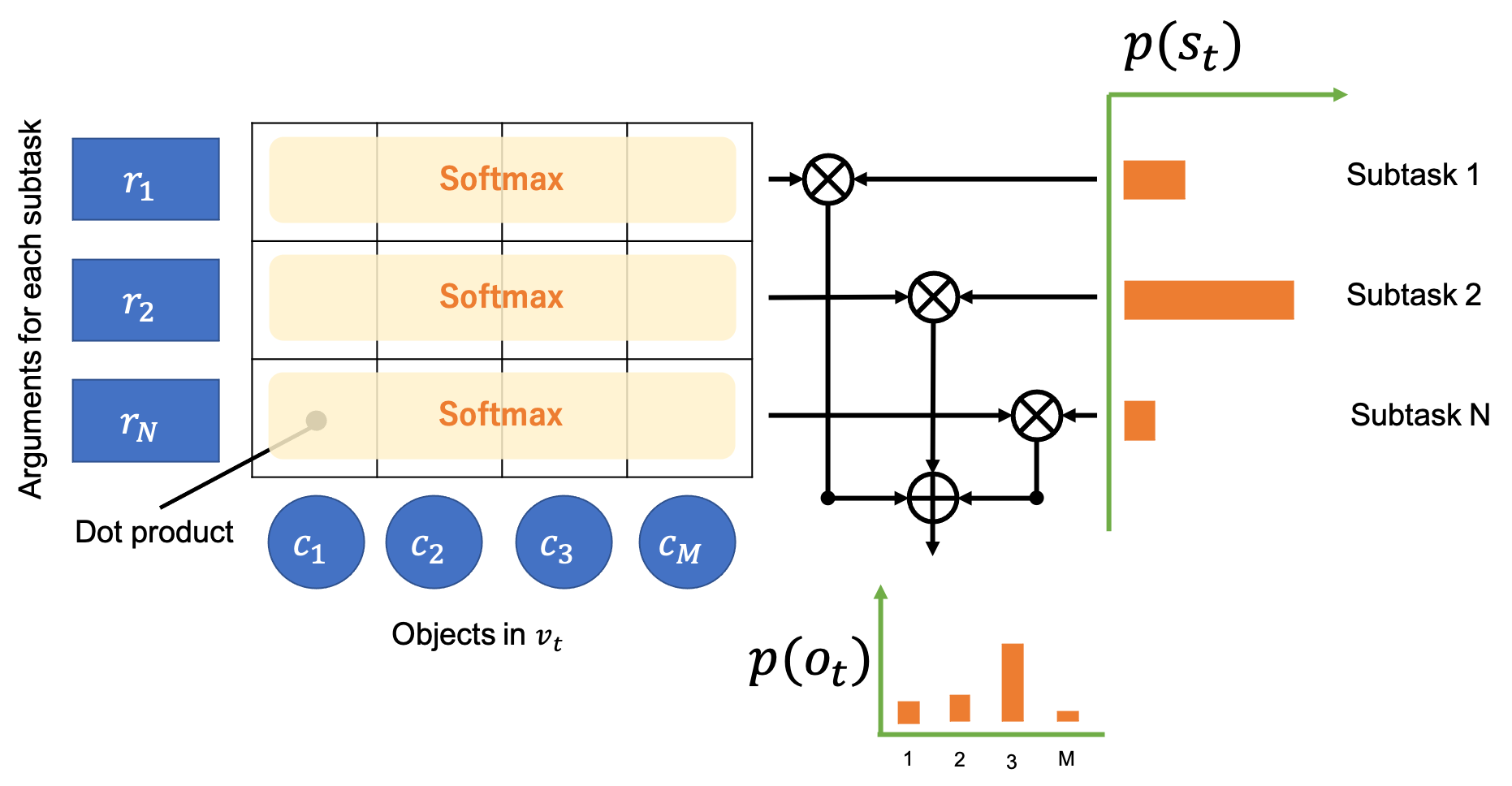}}
\caption{Detailed illustration of the object selector (\S\ref{sec:object-selector}).}
\label{fig:object_selector}
\end{figure*}

\subsection{Object Selector}
\label{sec:object-selector}
When the action $a_t$ is an interaction action such as Pickup or Slice, models need to select the object with a mask.
The object selector module outputs the mask of an selected object detected by MaskRCNN as follows:
\begin{align}
p(o_t) = & \sum_n p(s_t = n) {\rm Softmax}(E(c_{1:M}) E(r_n)^T)\\
m^* = & ~ {\rm argmax}_{o_t} p(o_t).
\end{align}
Then, the model outputs the mask $u_{m^*}$.
The overview of the object selector is shown in Figure \ref{fig:object_selector}.

\subsection{Progress Monitor}
Following Shridhar et al. \cite{ALFRED20}, our model learns the auxiliary task with the Progress Monitor, which monitors the progress of the task.
Specifically, from $h_t$ and $w_t$, we obtain normalized progress ($t/T$) and completed subtasks (number of accomplished subtasks divided by $N$) through independent linear layers.

\section{Experiments}
\subsection{Dataset}
We used the ALFRED dataset, in which roughly three annotators provided different language instructions for the final objective and each subtask for each demonstration played by skilled users of AI2-Thor \cite{ai2thor}.
ALFRED also provides the Planning Domain Definition Language (PDDL; \cite{mcdermott1998pddl}), which contains the high-level actions and their arguments. They are used to define the subtasks when creating the dataset.
In this study, we defined high-level actions and their arguments as the output of semantic understanding.
The number of training sets is 21,023. Since the test sets are not publicly available, we use the 820 validation sets for rooms that are seen during training, and the 821 validation sets for rooms that are not seen during training.
Note that the object to be selected in the validation set is an object that has never been seen during training, regardless of rooms.
Therefore, models need to be robust to unseen objects in both the validation sets.

\subsection{Training Details}
For NS-IF, we followed the hyperparameters proposed by Shridhar et al. \cite{ALFRED20}.
For RoBERTa-base, we used the implementation and default hyperparameters provided by Huggingface \cite{wolf2019huggingface}.
The hyperparameters for training NS-IF and RoBERTa-base are summarized in Table \ref{tb:hyperparam}.

\begin{table}[htbp]
    \centering
    \caption{Hyperparameters for training NS-IF and RoBERTa-base.}
    \begin{tabular}{ccc}
        \toprule
        \textbf{Hyperparameter} & ~ \textbf{NS-IF} ~ & \textbf{RoBERTa-base} \\\midrule
        Dropout & 0.3 & 0.1 \\
        Hidden size (encoder) & 100 & 768 \\
        Hidden size (decoder) & 512 & - \\
        Warmup ratio & 0.0 & 0.1 \\
        Optimizer & Adam \cite{kingma2014adam} & AdamW \cite{loshchilov2018decoupled} \\
        Learning rate & 1e-4 & 5e-5 \\
        Epoch & 20 & 5 \\
        Batch Size & 8 & 32 \\
        Adam $\epsilon$ & 1e-8 & 1e-8 \\
        Adam $\beta_1$ & 0.9 & 0.9 \\
        Adam $\beta_2$ & 0.999 & 0.999 \\
        Gradient Clipping & 0.0 & 1.0 \\
        \bottomrule
    \end{tabular}
    \label{tb:hyperparam}
\end{table}

\begin{table}[tbp]
\begin{center}
\caption{Success rate (\%) for each subtask in seen and unseen environments. The scores that take into account the number of actions required for success are given in parentheses. Higher is better. The best success rates among the models without oracle are \textbf{boldfaced}. The best success rates among all the models are \ul{underlined}.}
\begin{tabular}{l@{\hspace{4pt}}l@{\hspace{5pt}}r@{\hspace{5pt}}r@{\hspace{5pt}}r@{\hspace{5pt}}r}
& Model & \rotatebox{0}{Goto} & \rotatebox{0}{Pickup} & \rotatebox{0}{Slice} & \rotatebox{0}{Toggle}\\
\toprule
\multirow{5}{*}{\rotatebox{90}{Seen}} & S2S+PM \cite{ALFRED20} & - (51) & - (32) & - (25)  & - (100) \\ 
& S2S+PM (Reproduced) & 55 (46) & 37 (32) & 20 (15) & \textbf{\ul{100}} (100)\\
& MOCA \cite{Singh_2021_ICCV} & \textbf{\ul{67}} (54) & \textbf{64} (54) & 67 (50) & 95 (93) \\
& NS-IF & 43 (37) & \textbf{64} (58) & \textbf{71} (57) & 83 (83) \\\cmidrule{2-6}
& NS-IF (Oracle) & 43 (37) & \ul{69} (63) & \ul{73} (59) & \ul{100} (100) \\
\midrule
\multirow{5}{*}{\rotatebox{90}{Unseen}} & S2S+PM \cite{ALFRED20} & - (22) & - (21) & - (12) & - (32) \\
& S2S+PM (Reproduced) & 26 (15) & 14 (11) & 3 (3) & 34 (28) \\
& MOCA \cite{Singh_2021_ICCV} & \textbf{\ul{50}} (32) & \textbf{60} (44) & 68 (44) & 11 (10) \\
& NS-IF & 32 (19) & \textbf{60} (49) & \textbf{77} (66) & \textbf{43} (43) \\\cmidrule{2-6}
& NS-IF (Oracle) & 32 (19) & \ul{65} (53) & \ul{78} (53) & \ul{49} (49) \\
\bottomrule
\end{tabular}
\label{tb:subtask}
\end{center}
\end{table}

\begin{table}[ht]
    \centering
    \caption{Accuracy (\%) of semantic understanding (i.e., high-level action and argument prediction) for each subtask in seen and unseen environments.}
    \begin{tabular}{c|cccc|cccc}
    \toprule
    & \multicolumn{4}{c|}{High-level Action} & \multicolumn{4}{c}{Argument} \\\cmidrule{2-5}\cmidrule{6-9}
    & Goto & Pickup & Slice & Toggle & Goto & Pickup & Slice & Toggle \\\midrule
    Seen    & 99.40 & 99.01 & 91.39 & 97.87 & 71.58 & 89.68 & 92.72 & 76.60\\
    Unseen  & 99.22 & 98.84 & 97.14 & 99.42 & 73.73 & 89.78 & 96.19 & 64.74 \\
    \bottomrule
    \end{tabular}
    \label{tb:semantic-understanding}
\end{table}

\subsection{Main Results}
In this study, we evaluated the performance on each subtask, which is appropriate to assess the robustness to variations of objects and instructions in detail.
The baseline models are SEQ2SEQ+PM \cite{ALFRED20}, which uses only continuous representations in the computation process at each time, and MOCA \cite{Singh_2021_ICCV}, which factorizes the task into perception and policy.
Note that both the baselines are end-to-end neural models unlike our NS-IF.

We report the results in Table \ref{tb:subtask}.
The proposed NS-IF model improves the success rate especially in the tasks requiring object selection, such as Pickup, Slice and Toggle.
Notably, NS-IF improved the score on Slice in the Unseen environments from 3\% to 77\% compared to S2S+PM, and surpass MOCA.
The fact that only objects of unseen attributes need to be selected to accomplish the tasks in the test sets indicates that the proposed method is more robust to variations of objects on these subtasks than the baselines.

\begin{table*}[tbp]
\begin{center}
\caption{Three kinds of scores, (\Rnum{1}), (\Rnum{2}), and (\Rnum{3}), that reflect the robustness to variations of language instructions in the subtask evaluation.
These scores indicate the number of unique demonstrations where a model (\Rnum{1}) succeeds with all the language instructions, (\Rnum{2}) succeeds with at least one language instruction but fails with other paraphrased language instructions, or (\Rnum{3}) fails with all the language instructions.
Higher is better for (\Rnum{1}), and lower is better for (\Rnum{2}) and (\Rnum{3}). The best scores among the upper three models are \textbf{boldfaced}.}
\begin{tabular}{ll|r|r|r|r}
 & Model & Goto & Pickup & Slice & Toggle \\\toprule
\multirow{4}{*}{\rotatebox{90}{Seen}} & S2S+PM (Reproduced) & 315 / 240 / 239 & 105 / 52 / 202 & 7 / 5 / 29 & \textbf{29} / \textbf{0} / \textbf{0} \\
& MOCA \cite{Singh_2021_ICCV} & \textbf{386} / 281 / \textbf{131} & 184 / 90 / \textbf{86} & 24 / 8 / \textbf{9} & 26 / 2 / 1 \\
& NS-IF & 243 / \textbf{204} / 349 & \textbf{215} / \textbf{37} / 107 & \textbf{29} / \textbf{3} / \textbf{9} & 17 / 12 / \textbf{0} \\\cmidrule{2-6}
& NS-IF (Oracle) & 250 / 178 / 368 & 253 / 9 / 97 & 32 / 0 / 9 & 29 / 0 / 0 \\
\midrule
\multirow{4}{*}{\rotatebox{90}{Unseen}} & S2S+PM (Reproduced) & 147 / \textbf{99} / 513 & 42 / \textbf{21} / 281 & 1 / \textbf{0} / 31 & 13 / 10 / 30 \\
& MOCA \cite{Singh_2021_ICCV} & \textbf{216} / 307 / \textbf{233} & 155 / 84 / \textbf{103} & 18 / 6 / 7 & 4 / \textbf{6} / 44\\
& NS-IF & 168 / 145 / 441 & \textbf{182} / 36 / 122 & \textbf{24} / 1 / \textbf{6} & \textbf{19} / 9 / \textbf{25} \\\cmidrule{2-6}
& NS-IF (Oracle) & 165 / 89 / 502 & 218 / 12 / 113 & 25 / 0 / 7 & 28 / 0 / 25\\
& \multicolumn{5}{r}{(\Rnum{1}) $\uparrow$ / (\Rnum{2}) $\downarrow$ / (\Rnum{3}) $\downarrow$ }\\
\bottomrule
\end{tabular}
\label{tb:robustness-instruction}
\end{center}
\end{table*}

On the other hand, the S2S+PM model fails in many cases and does not generalize to unknown objects.
Moreover, the accuracy of S2S+PM is much lower in Unseen rooms than in Seen ones, which indicates that S2S+PM is less robust not only to unknown objects but also to the surrounding room environment.
By contrast, the difference in accuracy of NS-IF between Seen and Unseen is small, indicating that the proposed model is relatively robust to unknown rooms.
This may be related to the fact that the output of ResNet is sensitive to the scenery of the room, while the output of MaskRCNN is not.
The failed cases of NS-IF in Pickup and Slice are caused by the failure to predict the action $a_t$, or failure to find the object in drawers or refrigerators after opening them.

There are still some shortcomings in the proposed model.
There was little improvement in the Goto subtask.
It may be necessary to predict the bird's eye view from the first person perspective, or the destination based on the objects that are visible at each time step.
In addition, the accuracy of other subtasks (PutObject, etc.) that require specifying the location of the object has not yet been improved.
This is because the pre-trained MaskRCNN used in this study has not been trained to detect the location of the object.

\subsection{Performance of Semantic Understanding}
To investigate the cause of the performance gap between NS-IF and its oracle, we evaluated the performance of the semantic understanding module for each subtask.
The results are given by Table \ref{tb:semantic-understanding}.

The accuracies of high-level action prediction are 91 $\sim$ 99 \%.
Whereas, the accuracies of argument prediction are 64 $\sim$ 96\%.
This may be because the number of classes of arguments are 81, while that of high-level actions is eight.

For the Toggle subtask, the accuracy of argument prediction is lower than 80\%.
This error might primarily cause the drop of the success rate in Toggle in Table \ref{tb:subtask} from NS-IF to NS-IF (Oracle).
Thus, improving the accuracy of argument prediction would close the gap.

In contrast, despite of the error of argument prediction in Goto as seen in Table \ref{tb:semantic-understanding}, the success rates of NS-IF and its oracle in Table \ref{tb:subtask} were almost the same.
This observation implies that our NS-IF failed to fully utilize the arguments to perform the Goto subtasks.
Mitigating this failure is future work.

\section{Analysis: Evaluating the Robustness to Variations of Language Instructions}
The robustness of models to variations of language instructions can be evaluated by seeing whether the predictions remains correct even if the given language instructions are replaced by paraphrases (e.g., Figure \ref{fig:pickup}) under the same conditions of the other variables such as the room environment and the action sequence to accomplish the task.

The results are shown in Table \ref{tb:robustness-instruction}.
The reported scores show that the proposed model increased the overall accuracy while improving the robustness to variations of language instructions compared to S2S+PM.
The numbers of demonstrations corresponding to (\Rnum{1}), ``succeeds with all the language instructions'', for NS-IF were superior to the baselines for Pickup, Slice, and Toggle in unseen environments, which indicates that NS-IF is the most robust to paraphrased language instructions.
Using oracle information further increased the robustness.

The cases that fall into the category (\Rnum{3}), ``fails with all the language instructions", are considered to result from causes unrelated to the lack of the robustness to variations of language instructions.
These failures are, for example, caused by the failure to select an object in a drawer or a refrigerator after opening them.

\section{Related Work}

\subsection{Neuro-Symbolic Method}
In the visual question answering (VQA) task, Yi et al. \cite{yi-etal-2018-neural-symbolic} proposed neural-symbolic VQA, where the answer is obtained by executing a set of programs obtained by semantic parsing from the question against a structural symbolic representation obtained from the image using MaskRCNN \cite{maskrcnn}.
Reasoning on a symbolic space has several advantages such as (1) allowing more complex reasoning, (2) better data and memory efficiency, and (3) more transparency, making the machine's decisions easier for humans to interpret.
In the VQA task, several similar methods have been proposed.
Neuro-Symbolic Concept Learner \cite{mao-et-al-2019-neuro-symbolic} uses unsupervised learning to extract the representation of each object from the image and analyze the semantics of the questions.
Neural State Machine \cite{hudson2019abstraction} predicts a scene graph including not only the attributes of each object but also the relationships between objects to enable more complex reasoning on the image.
However, they are different from our study in that they all deal with static images and the final output is only the answer.
Neuro-symbolic methods were also applied to the video question answering task, where a video, rather than a static image, is used as input to answer the question \cite{yi-etal-2020-clevrer}. However, here too, the final output is only the answer to the question.

\subsection{Embodied Vision-and-Language Task}
Tasks that require an agent to move or perform other actions in an environment using vision and language inputs have attracted much attention in recent years.
In the room-to-room dataset \cite{Anderson_2018_CVPR}, a Vision-and-Language Navigation task was proposed to follow language instructions to reach a destination.
In both the embodied question answering \cite{Das_2018_CVPR} and interactive question answering \cite{gordon2018iqa} tasks, agents need to obtain information through movement in an environment and answer questions, and the success or failure is determined by only the final output answer.
In contrast to these tasks, ALFRED \cite{ALFRED20} aims to accomplish a task that involves moving, manipulating objects, and changing states of objects.

\section{Conclusion}
We proposed a neuro-symbolic method to improve the robustness to variations of objects and language instructions for interactive instruction following.
In addition, we introduced the subtask updater that allows the model to recognize which subtask is being solved at each time step.
Our experiments showed that the proposed method significantly improved the success rate in the subtask requiring object selection, while the error propagated from the semantic understanding module degraded the performance.
The experimental results suggest that the proposed model is robust to variations of objects.
The analysis showed that the robustness to variations of language instructions was improved by our model.

ALFRED contains the ground truth output of semantic understanding and the prior knowledge of which subtask was being solved at each step, so it was possible to use them in training.
It should be noted that the cost of annotations of them can not be ignored for other datasets or tasks.
If the cost is impractical, it may be possible to solve the problem by unsupervised learning, as in NS-CL \cite{mao-et-al-2019-neuro-symbolic}.
Whereas, for training MaskRCNN, annotation is not necessary because the mask and class information of the object can be easily obtained from AI2-Thor.
Therefore, whether annotation of mask and class is necessary or not depends on how well an object detection model trained on artificial data obtained from simulated environments generalizes to real world data.
Future work includes learning subtack updater to enable evaluation on the whole task.

\section*{Acknowledgements}
This work benefited from informal discussion with Shuhei Kurita. The authors also thank the anonymous reviewers for their valuable comments.
This work was supported by JSPS KAKENHI Grant Number 22J13751.

%
%
%
\bibliographystyle{splncs04}
%
\bibliography{anthology,custom}
\end{document}